\documentclass[a4paper]{article}
\usepackage{todonotes}
\usepackage{INTERSPEECH2020}
\usepackage{multirow,url}

\title{The Zero Resource Speech Challenge 2020:\\ Discovering discrete subword and word units}

\name{Ewan Dunbar$^{1,2*}$,
      Julien Karadayi$^2$,
      Mathieu Bernard$^{2}$,
      Xuan-Nga Cao$^{2}$,
      Robin Algayres$^2$,
      Lucas Ondel$^3$,
      Laurent Besacier$^4$,
      Sakriani Sakti$^{5,6}$,
      Emmanuel Dupoux$^{2,7}$}
\address{
  $^1$Universit{\'{e}} de Paris, LLF, CNRS, Paris, France\\
  $^2$Cognitive Machine Learning (ENS - CNRS - EHESS - INRIA - PSL Research University), France\\
  $^3$Department of Computer Graphics and Multimedia, Brno Univ. of Technology, Czech Republic\\
  $^4$Laboratoire d'Informatique de Grenoble, \'equipe GETALP (Universit\'e Grenoble Alpes), France\\
  $^5$Nara Institute of Science and Technology\\
  $^6$RIKEN Center for Advanced Intelligence Project, Japan
  $^7$Facebook A.I. Research, Paris, France
}
\email{$^*$ewan.dunbar@utoronto.ca}

\begin{document}
\maketitle

\begin{abstract}
We present the Zero Resource Speech Challenge 2020, which aims at learning speech representations from raw audio signals without any labels. It combines the data sets and metrics from two previous benchmarks (2017 and 2019) and features two tasks which tap into two levels of speech representation. The first task is to discover low bit-rate subword representations that optimize the quality of speech synthesis; the second one is to discover word-like units from unsegmented raw speech. We present the results of the twenty submitted models and discuss the implications of the main findings for unsupervised speech learning. 
\end{abstract}

\section{Introduction}

Current speech technology depends heavily on the availability of textual resources. On the other hand, humans learn the sounds and vocabulary of their first language long before they learn to read or write, discovering some kind of linguistic units or representations in their language (typically thought to be phoneme- or word-like), and the equivalent of an acoustic model, a language model, and a speech synthesizer. That humans succeed without textual resources suggests that there may be another approach. Developing technology to learn useful speech representations in an unsupervised way would be useful for the thousands of so-called low-resource languages, which lack the textual resources and/or expertise required to build traditional speech processing systems. 

\begin{figure}[h]
  \centering
  \includegraphics[width=.8\linewidth]{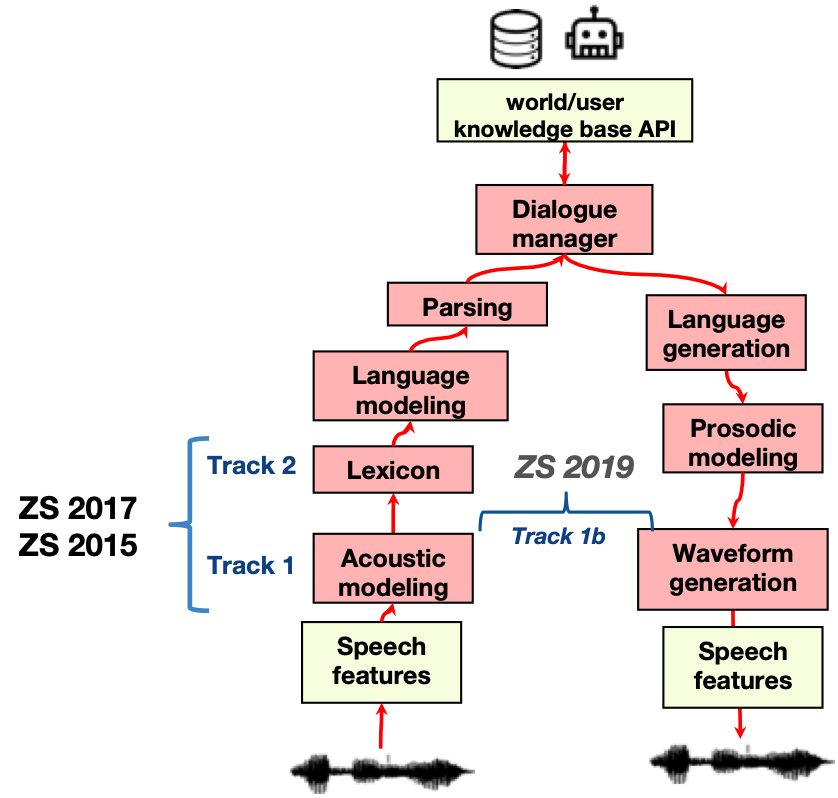}
  \caption{Schematic diagram of the Zero Resource challenge series. The long term aim is to learn an entire spoken dialogue stack without recourse to any textual resources, mimicking the way human children learn their native languages. }
  \label{fig:diagram-challenge}
\vspace{-2em}
\end{figure}

The Zero Resource Speech Challenge series\footnote{Detailed results of all metrics, and audio samples for unit discovery/synthesis systems, are provided on the leaderboard at \url{http://www.zerospeech.com/}.} \cite{versteegh2016,dunbar2017,dunbar2019} aims to push the envelope in unsupervised speech modelling, by taking the radical stance of trying to learn the full speech processing stack without any textual resources. Here, we reopen two previous benchmarks with a focus on discovering discrete representations from raw audio at two linguistic levels. The first focuses on the phonological or sub-word level. The goal is to learn discrete (low bitrate) speech units, which encode meaningful linguistic invariants, and which are useful for doing speech synthesis. This is a reopening of the 2019 ``TTS without T'' ZeroSpeech benchmark \cite{dunbar2019}  (track 1b in Figure \ref{fig:diagram-challenge}). The second focuses on the word level. The  goal is to discover word-like units for the purpose of segmenting continuous speech. This is a reopening of the 2017 ``spoken term discovery''   ZeroSpeech Benchmark  \cite{dunbar2017} (track 2 in Figure \ref{fig:diagram-challenge}). As before, we rely exclusively on freely accessible software and data sets. 


Discrete units, such as words and phonemes, form the basis of every modern speech 
technology system at some level. 
One useful feature of discrete representations is that they remove linguistically irrelevant information from the signal, and represent continuous speech in a highly compact 
format. 
For example,  \cite{DBLP:conf/iclr/BaevskiSA20} perform unsupervised representation learning, and show that, up to a certain point, discrete representations are more useful than continuous ones as the input for training a phone recognizer.
Here, we ask participants to discover their own discrete units and analyze them in terms of how well they capture relevant linguistic contrasts, as indicated by the gold phoneme- and  word-level transcriptions.

\section{Data sets, metrics and baselines}
\subsection{Unsupervised unit discovery for speech synthesis}

\textit{Task.} The problem of learning speech units useful for doing speech synthesis can be seen in terms of an encoder--decoder architecture. The encoder takes as input raw audio and turns it into a sequence of speaker-invariant discrete units (``pseudo-text''). The decoder takes these units as input and generates a waveform corresponding to the same linguistic content uttered in another voice. Requiring synthesis is in a new voice allows us to exclude trivial solutions where the audio is returned unchanged. We measure the \textbf{synthesis quality} of the output, as well as the \textbf{unit quality} and the \textbf{bitrate} of the pseudo-text.

\textit{Data sets.} We use the same setting and datasets as in \cite{dunbar2019}, with two languages, the development language (English) and the surprise language (Indonesian). Participants are instructed to treat these languages as low-resource and refrain from using other labelled data. Metrics are provided only for the development language, and results for the surprise language must be submitted through the challenge website to be evaluated (maximum two submissions per research group).  Three unlabelled data sets are provided for each language.
The \textit{Train Voice} data set contains either one talker (surprise) or two  (development), and is intended for building an acoustic model of the target voice for speech synthesis (between 1h30 and 2h40 of data per voice). The \textit{Train Unit Discovery} data set contains read speech from multiple speakers, with around ten minutes of speech from each speaker, for a total of 15h in each language. These are intended for the discovery of speaker-independent acoustic units. 
The \textit{Test} data set contains new utterances from unseen speakers.



\emph{Synthesis quality} is measured using human evaluation, taking three measures. To evaluate the \emph{comprehensibility} of the synthesis, the evaluators were asked to orthographically transcribe the synthesized sentence. Each transcription was compared with the gold transcription using the Levenshtein distance, yielding a character error rate (CER).  The overall \emph{naturalness} of the synthesis was assessed on a 1 to 5 scale, yielding a mean opinion score (\textit{MOS}), where 5 is the most natural. Finally, the \textit{similarity} of the voice to the target voice was assessed using a 1 to 5 scale, with each synthesized sentence compared to a reference sentence uttered by the target voice, with 5 being the most similar to the target. Each evaluator performed the evaluation tasks in the same order (comprehensibility, naturalness, similarity), with the overall evaluation lasting about one hour. We recruited evaluators for English using the Amazon Mechanical Turk platform and, for Indonesian, through universities and research institutes in Indonesia. All were paid the equivalent of 10 USD. Catch trials (novel natural recordings by the target voice) ensured that participants were on task: only data from participants with $<$0.80 CER on catch trials was retained (English: 35; Indonesian: 68).

\textit{Embedding bitrate and quality}. Participants submit encodings for each test file. The submitted encodings are sequences of vectors. These vectors are assumed to be quantized. We calculate the \emph{bitrate} of the encoding by, first, constructing a dictionary of all distinct vectors over the entire test set. The test set is seen  as a sequence $U=[s_1,...,s_n]$ of $n$ symbols. The bitrate is calculated as $n \sum_{i=1}^{n}{\frac{p(s_i)log_{2}p(s_i)}{D}}$, where $p(s_i)$ is the relative frequency of symbol $s_i$ in $U$, and $D$ the total duration of $U$ in seconds.
The \emph{unit quality}  is evaluated with the \textit{ABX phone discriminability score}, as in previous Zero Resource challenges \cite{versteegh_2016,dunbar2017}. The ABX discriminability, for example, between [aba] and [apa], is defined as the probability that the representations of $A$ and $X$ are more similar than representations of $B$ and $X$, over all triplets of tokens such that $A$ and $X$ are tokens of [aba], and $B$ a token of [apa] (or vice versa), and such that $X$ is uttered by a different speaker than $A$ and $B$. The global ABX phone discriminability score aggregates over the entire set of minimal triphone pairs such as [aba]--[apa] to be found in the test set. The choice of the appropriate distance measure is up to the researcher. As in previous challenges, we provide a default distance, the average frame-wise angle (arc cosine of the normalized dot product) between the embeddings of the tokens along a DTW-realigned path, and also make available an equivalent distance making use of frame-wise symmetrised KL-divergences, rather than angles, as well as a Levenshtein (edit) distance measure. We cite ABX scores as error rates (0\% for the gold transcription, 50\% being chance). Each of the items compared in the ABX task is a triphone ([izi]-[idi], and so on), extracted from the test corpus.
Each triphone item is a short chunk of extracted audio, to be decoded by the systems.\footnote{This differs from previous challenges. In previous challenges, longer audio files were provided for decoding, from which the representations of triphones were extracted after the fact using time stamps. In the 2019/2020 edition, triphones are pre-extracted, to allow for systems without fixed frame rates.}


\textit{Toplines and baselines}. A baseline system is provided, consisting of a pipeline with a nonparametric Bayesian acoustic unit discovery system  \cite{DBLP:conf/sltu/OndelBC16,ondel2018bayesian}, and a parametric speech synthesizer based on Merlin \cite{DBLP:conf/ssw/WuWK16}. As linguistic features, we use contextual information (leading and preceding phones, number of preceding and following phones in current sentence), but no features related to prosody, articulatory features (vowel, nasal, and so on), or part-of-speech (noun, verb, adjective, and so on). The baseline system is made available in a container. A supervised topline system is also provided, consisting of a phone recognizer trained using Kaldi \cite{Povey_ASRU2011_2011} on the original transcriptions. The acoustic model is a tristate triphone model with 15000 Gaussian mixtures. The language model is a trigram phone-level language model.\footnote{A word-level language model gives better performance, but we use a phone-level language model in the interest of giving a fair comparison with the subword unit discovery systems asked for in the challenge.} Output is piped to the TTS system, which is also trained on the gold labels.


\begin{table}[h]
  \caption{Submissions to the \textbf{unit discovery/synthesis} track.}
  \label{tab:systems}
  \centering
  \setlength\tabcolsep{5 pt}
  \begin{tabular}{p{1.5cm}|p{2.4cm} p{2.46cm}}
 & \textbf{Encoder/Decoder}  &\textbf{Generation}   \\
    \hline
     \textbf{MC} \cite{chen20} (Sheffield) & Disentangled discrete AEs & Wavenet \\ 
     \textbf{TM} \cite{morita20} (Kyoto)     & ABCD-VAE & Neural source-filter\\
     \textbf{BN} \cite{niekerk20} (SU)  & VQVAE\,(1),  VQCPC\,(2)  & WaveRNN \\ 
     \textbf{PL} \cite{lumbantobing20} (Nagoya)  & CycleVQVAE\,(1), Cycle-VAE\,(2) & Wavenet \\ 
     \textbf{BY} (Brno)     & Subspace HMM + AUD + Baseline & Baseline \\ 
     \textbf{MK} \cite{pandia20} (IIT)           & CV, VC transients  & Waveglow \\ 
     \textbf{AT} \cite{tjandra20} (NAIST)      & VQVAE + transformer & Griffin-Lim (2) \\ 
     \textbf{BG} \cite{gundogdu20} (Bo\u gazi\c ci)    & Correspondence rec. sparse AE & Baseline \\
     \textbf{WH} (Tokyo IT)    & Hierarchical VQ-VAE & MelGAN \\
    \hline
  \end{tabular}
\end{table}


\subsection{Spoken term discovery \& segmentation}

\textit{Task.} The goal of spoken term discovery is to find words in the speech stream---just as the infant learns the words of its language by listening. The input is a series of speech features. The output is a set of boundaries delimiting the start and end of proposed word tokens discovered in the speech, and category labels indicating proposed word types. These boundaries may, but need not, constitute an exhaustive parse of the speech. The evaluation we apply is a set of scores measuring different aspects of the alignment with the words in the gold-standard transcription.  As is customary in the field of word segmentation, we do not provide a separate test set for this track; we rely on the surprise languages to assess possible hyperparameter overfitting.
Two submissions per research group are allowed.

\textit{Data sets}.
The \emph{development data} and \emph{surprise data} are the same as in \cite{dunbar2017} (see \cite{sakti2008developmentasr,DBLP:conf/asru/TjandraS017}). The development data consists of corpora from three languages (English, French and Mandarin). Each corpus comes with software that performs the evaluation. Challenge participants are encouraged to use these resources to tune their hyperparameters using a cross-validation approach to maximize generalizability. The participants then must submit their systems and their output on all  the data  sets for independent evaluation (run automatically upon submission). The surprise data consists of corpora from two additional languages (German and Wolof), which are provided with no additional resources. 

The amount of data in the training part of the development data sets varies from 2.5 to 45 hours, to ensure that systems can work both with limited data and with larger data sets. The statistics of the two surprise languages fall between these two extremes.  The distribution of speakers in the training sets is shaped to reflect what is typically found in natural language acquisition settings: there is a ``family''---a small number of speakers (between four and ten) who make up a large proportion of the total speech---and a set of ``outsiders''---a larger number of speakers that each appear in smaller proportions (ten minutes each). The test sets consist of many short files, are organized into subsets of differing length (1s, 10s and 120s).

The English and French corpora were taken from LibriVox audio books\footnote{http://librivox.org/} and phone force-aligned using Kaldi \cite{Povey_ASRU2011_2011}. The Mandarin corpus is described in \cite{mandarinpaper}, force-aligned using Kaldi. The  German corpus was taken from LibriVox and force-aligned using Kaldi as well. The Wolof corpus is described in \cite{wolofpaper}.


\textit{Evaluation metrics}. ``Spoken term discovery'' is a complex task with a number of sub-goals, which can be evaluated separately. The first sub-goal is to do good \emph{matching:} deciding whether any two given speech fragments are instances of the same sequence of phoneme, and attempting to find as many matches as possible. The \emph{quality of matches} is evaluated based on how similar fragments matched by the system are---we use the average normalized edit distance (\textbf{NED}) between the gold phoneme sequences, over all pairs matched by the system---the \emph{quantity of matches} can be evaluated by measuring the proportion of the corpus covered by matched pairs (\textbf{coverage}).  

The second sub-goal is to construct a lexicon. This amounts to \emph{clustering} the discovered matched pairs. The \emph{intrinsic quality} of the lexicon is evaluated based on how consistent the items clustered together are with regard to the sequences of gold phonemes they correspond to.  The \textbf{Grouping} scores (precision, recall and F-score) evaluate the purity and inverse fragmentation of the clusters in a pairwise fashion (see \cite{dunbar2017} for a formal definition). 
The \emph{extrinsic quality} can be measured with respect to how well the clusters match the gold-standard lexicon of word types. \textbf{Type} scores (precision, recall and F-score) measure the correspondence between the discovered clusters and the gold lexicon. Type precision is the probability that discovered types belong to the gold set of types (real words), type recall is the probability that gold types are discovered. We restrict both sets to words between three and twenty phones long.

\begin{figure*}[h]
  \centering
 
  \begin{tabular}{p{2in}p{2.25in}|p{2.25in}}
   \includegraphics[width=\linewidth]{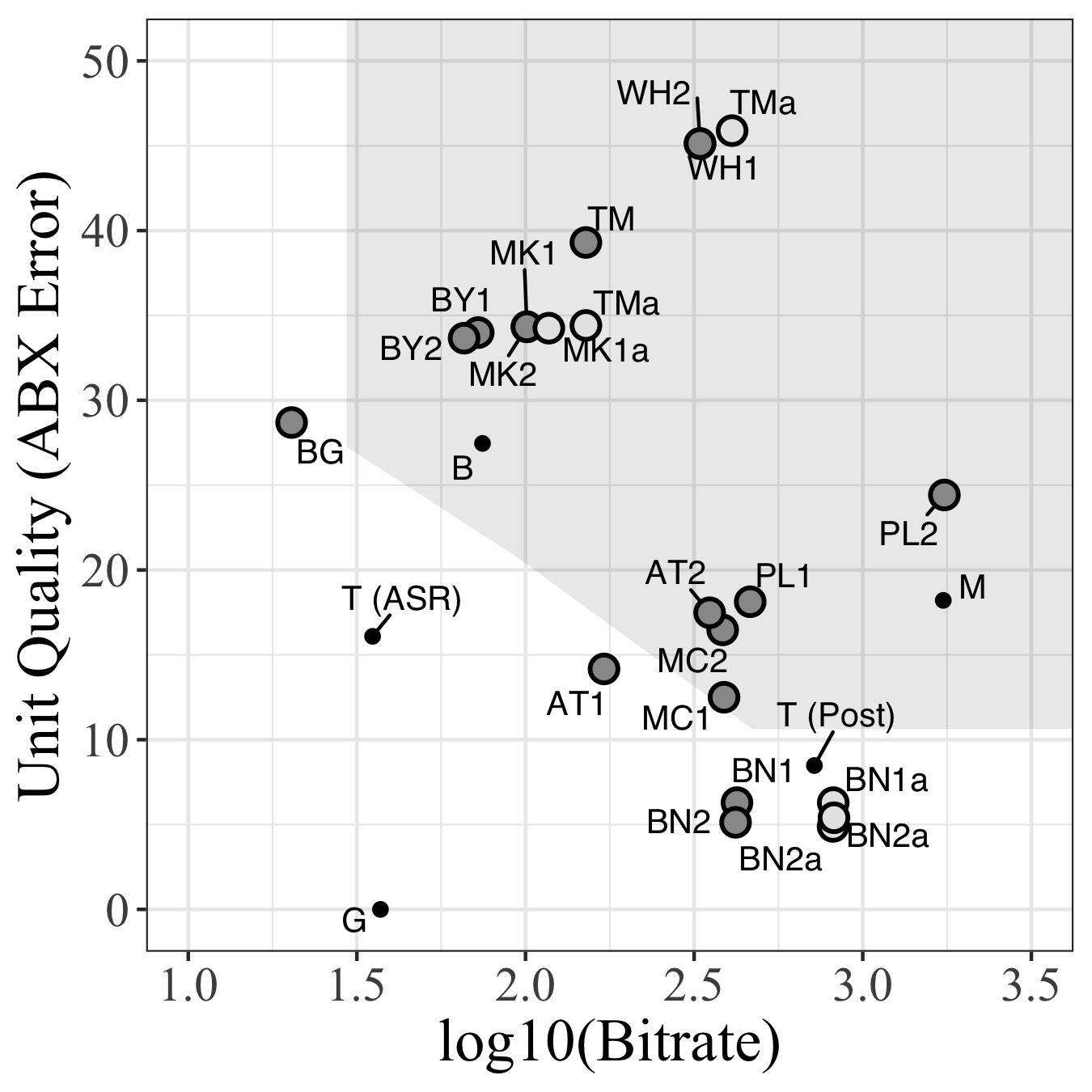} &   \includegraphics[width=.8888\linewidth]{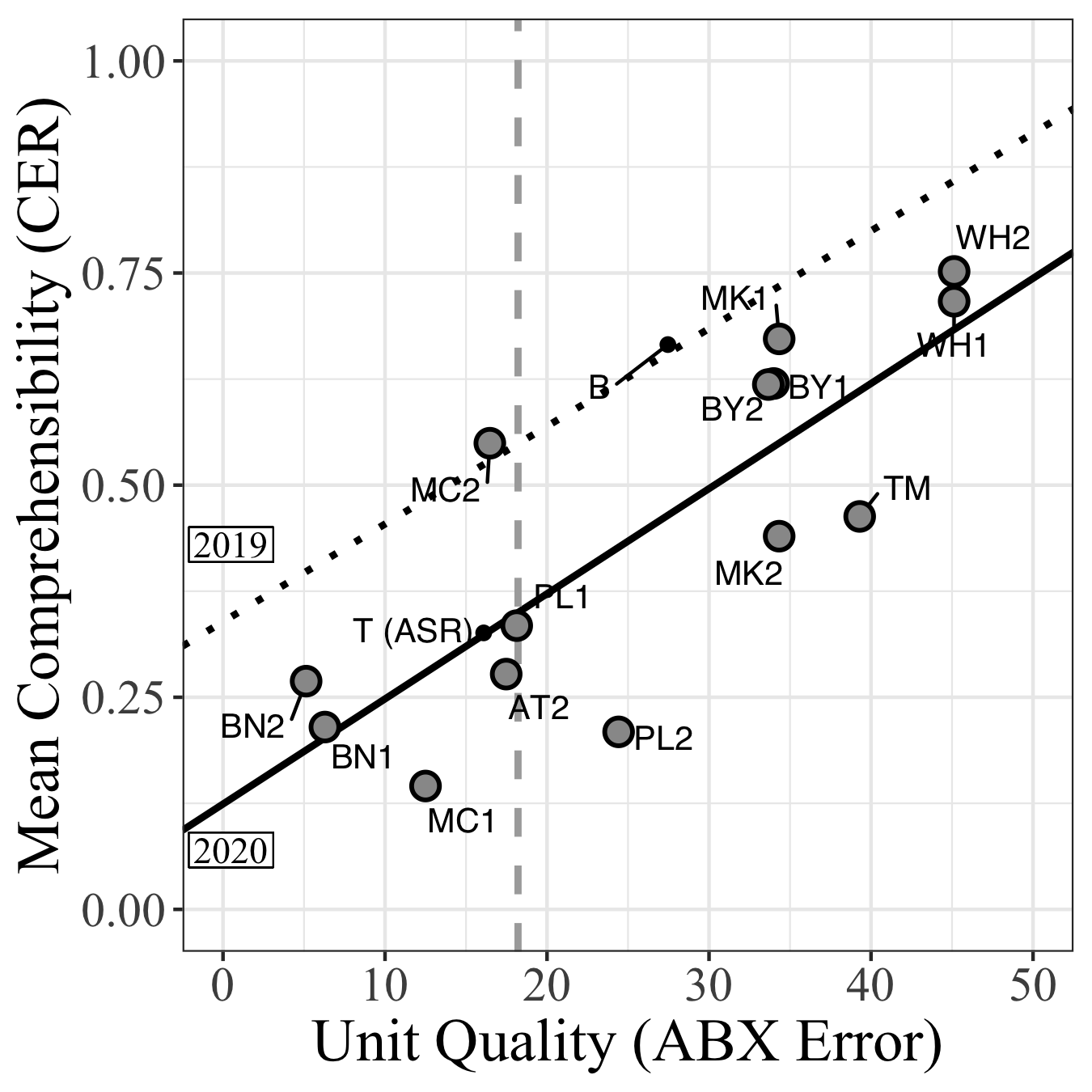}& ~~~~\includegraphics[width=2in]{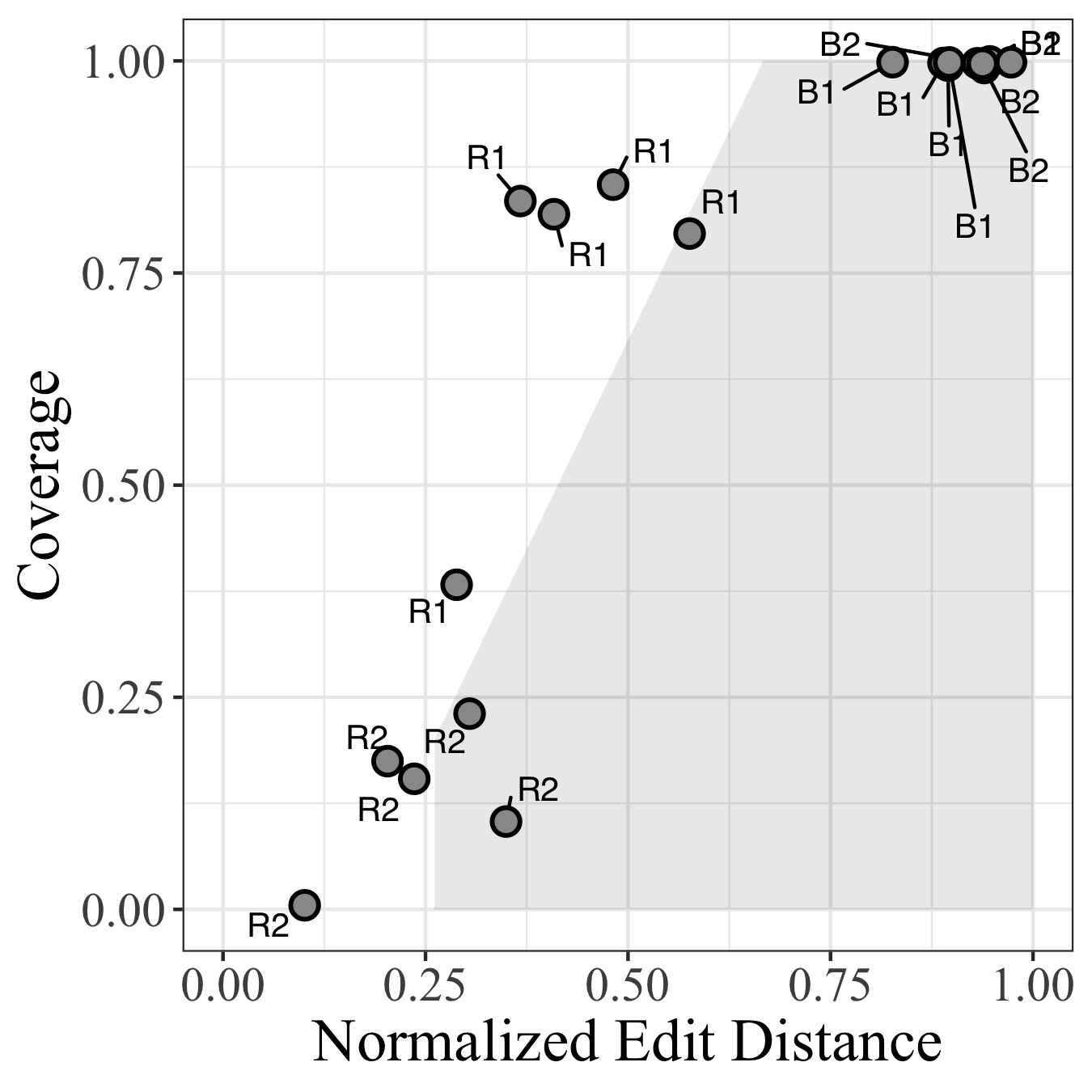} \\
 (a) & (b) & (e) \\
   \includegraphics[width=\linewidth]{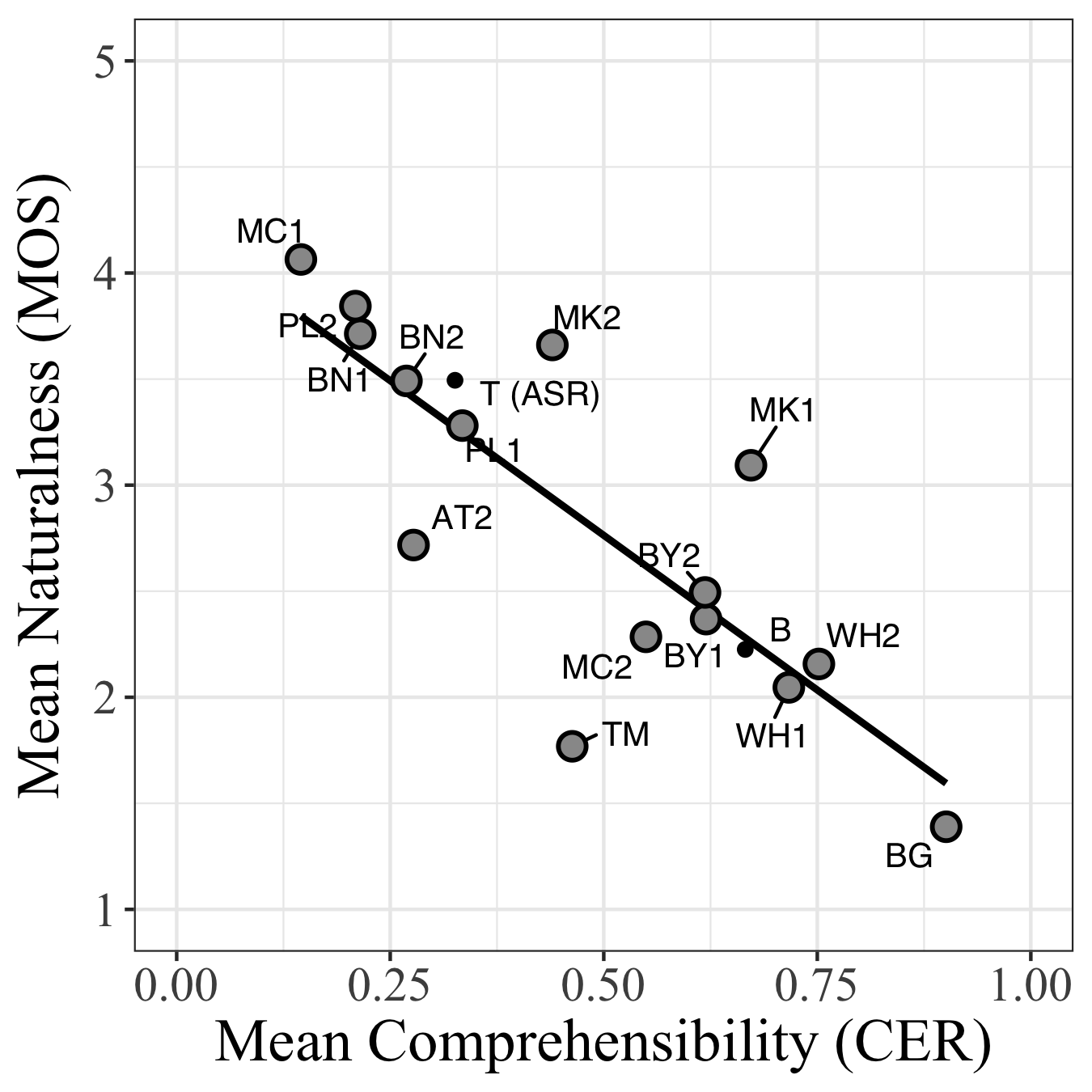} &   \includegraphics[width=.8888\linewidth]{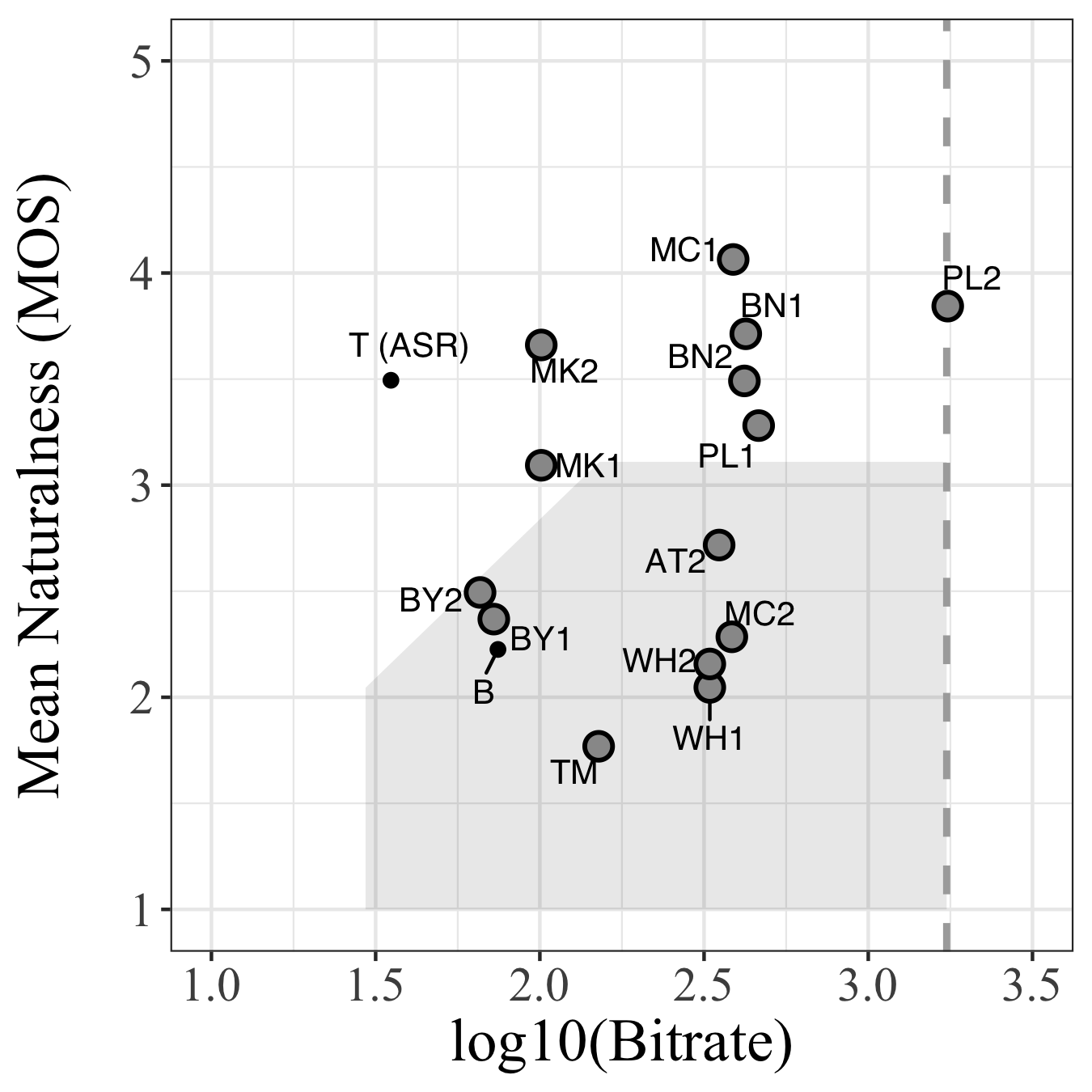} & ~~~~\includegraphics[width=2in]{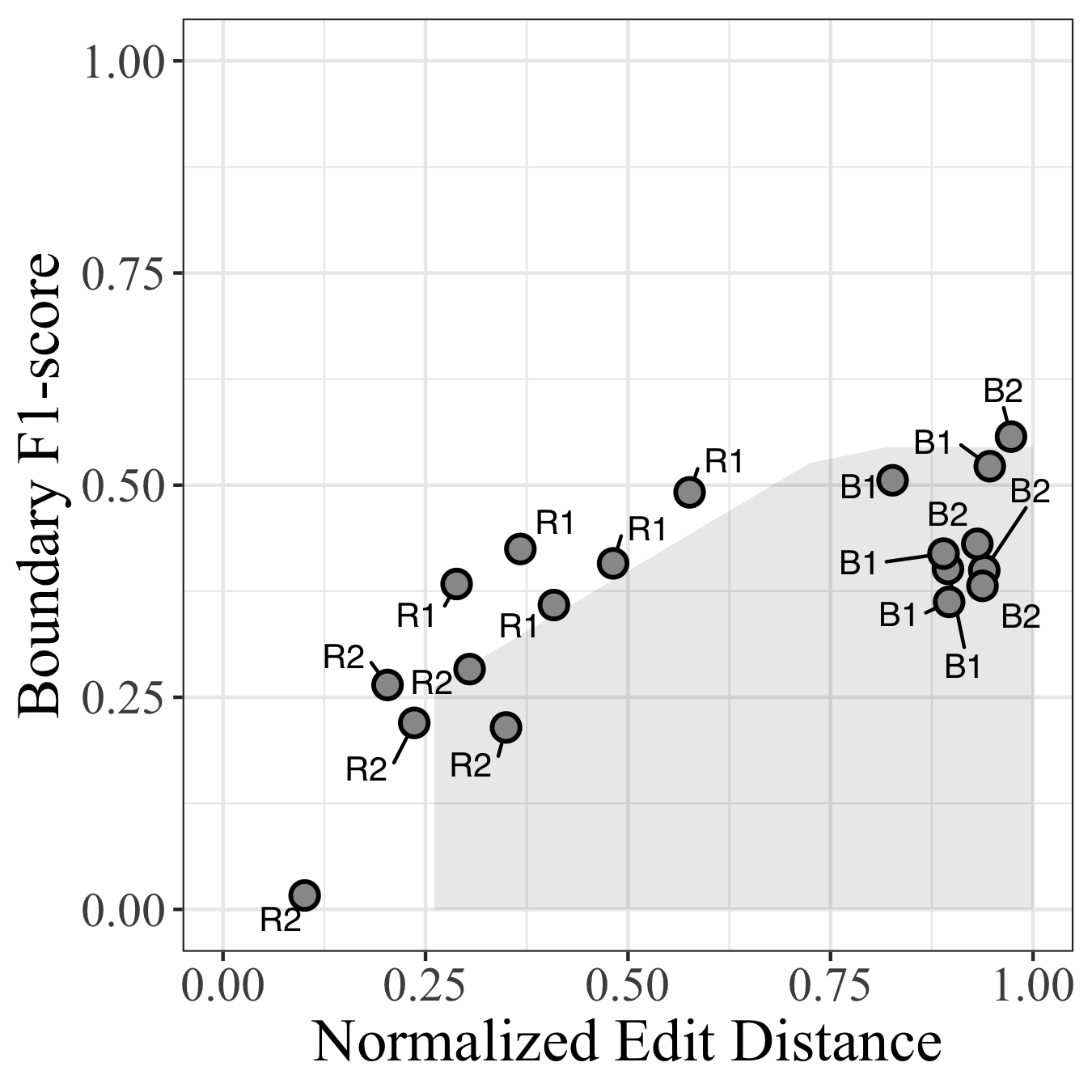} \\
 (c) & (d) & (f) \\
  \end{tabular}

  \caption{\textbf{(a)} ABX error (lower better) as function of bitrate for \textbf{unit discovery/synthesis}. \textbf{(b)} Character error rate (CER: lower is better) as a function of ABX error. Vertical dashed line: ABX error for MFCCs. Sloped lines are linear regressions for 2019 submissions (dotted) and for 2020 submissions (solid), showing global increase in decoding quality. \textbf{(c)} Mean opinion score (MOS: higher is better) as a function of CER. Line is linear regression. \textbf{(d)} MOS as function of bitrate.  Vertical dashed line is MFCC bitrate. \textbf{Unit discovery/synthesis}  results presented on surprise language only. Reference scores plotted as \textbf{G} for gold transcriptions; \textbf{M} for MFCC features; \textbf{B} for baseline system; \textbf{T (Post)} for posteriorgrams from the topline system; and \textbf{T (ASR)} for discrete decoding from the topline. Edge of grey regions in \textbf{(a)} and \textbf{(d)} represents 2019 state of the art on the tradeoff. Labels are 2020 submissions. \textbf{(e)}. Coverage (higher is better) as a function of normalized edit distance (NED: lower is better) for \textbf{spoken term discovery/segmentation} submissions. \textbf{(f)} Boundary F-score (higher is better) as a function of NED for  submissions. Edge of grey regions in \textbf{(e)} and \textbf{(f)} represents 2017 state of the art on the tradeoff, labels are 2020 submissions, and multiple points per system are different languages. Clustering-oriented algorithms have low NED, while segmentation-oriented algorithms have high coverage and boundary F-scores.}
  \label{fig:results}
\vspace{-2em}
\end{figure*}

The third sub-goal is to do \emph{accurate word segmentation.} The \textbf{Token} scores (precision, recall and F-scores) evaluate the quality of the discovered fragment tokens compared to the gold tokens, and the \textbf{Boundary} scores (precision, recall and F-scores) the quality of the discovered boundaries.

By setting out three different types of criteria, the intention is to be open to various types of ``spoken term discovery'' systems, all of which in some sense ``find words.'' The result is that we do three (non-independent) types of evaluations. All of these evaluations are done at the level of the phonemes: using the aligned phoneme transcription, we convert any discovered fragment of speech into its transcribed string.  If the left or right edge of the fragment contains part of a phoneme, that phoneme is included in the transcription if it corresponds to  more than 30ms or more than 50\% of its duration.

\textit{Baselines and toplines}. The baseline was computed using \cite{aren}, which does pair-matching using locally sensitive hashing applied to PLP features and then groups pairs using graph clustering. The parameters stayed the same across all languages, except that the dynamic time warping threshhold was increased for Mandarin (to 0.90, rather than 0.88), in order to obtain a NED value similar to that of other languages. The topline system was an exhaustive-parsing word segmentation model based on the textual transcriptions (a unigram grammar trained  in the adaptor grammar framework: \cite{adaptorgrammar}).

\section{Models and selected results}

\subsection{Unsupervised unit discovery for speech synthesis}

Sixteen systems from nine teams were submitted, summarized in Table \ref{tab:systems}. Two systems, \textbf{AT-1} and \textbf{BG}, are excluded from analysis of the synthesis evaluation due to declared issues with the submissions. Relatively few systems were submitted in the ``low bitrate'' range (near the bitrate of the annotation). Nevertheless, the systems submitted this year, which have shifted towards higher bitrates and end-to-end systems mostly based on discrete autoencoders, have all done more with less. Figure \ref{fig:results}a shows (for the surprise language) the improvements in embeddings with respect to the previous year: the edge of the grey zone shows the   empirical tradeoff previously observed between \textbf{unit quality} and \textbf{bitrate}. Many of this year's systems improve reach lower ABX error rates at a given bitrate. Figure \ref{fig:results}b shows that improvements have also been made in \textbf{decoding,} with overall more comprehensible synthesis, regardless of unit quality (the 2019 systems are represented by the dotted line of best fit, while the solid line is fit through the current submissions). And, while the comprehensibility measure is largely correlated with the overall synthesis naturalness evaluations (MOS), Figure \ref{fig:results}c shows that certain systems are reported to sound particularly natural, beyond just their comprehensibility (notably the two \textbf{MK} systems). This is presumably due to a improvement in \textbf{waveform generation.} Figure \ref{fig:results}d shows the combined effect of these improvements in \textbf{unit quality}, \textbf{decoding,} and \textbf{waveform generation,} showing major improvements on the tradeoff between synthesis quality and bitrate.










\subsection{Spoken term discovery \& segmentation}

%

Two teams, indicated in Figure \ref{fig:results} as \textbf{B} \cite{bhati20} (JHU) and \textbf{R} \cite{rasanen20} (Tampere), submitted two systems each. The edge of the grey region in Figure \ref{fig:results}e  shows the  empirical tradeoff previously observed between having \textbf{high quality matching} (low NED) and \textbf{exhaustively analysing} the corpus (high coverage). Systems \textbf{R1} and \textbf{R2}, which employ probabilistic dynamic time warping,  both clearly improve on the tradeoff, with \textbf{R1} privileging exhaustiveness and \textbf{R2} match quality. Figure \ref{fig:results}f  shows the  empirical tradeoff between high quality matching and \textbf{accurate word segmentation}. Systems \textbf{R1} and \textbf{R2} again show improvement. Systems \textbf{B1} and \textbf{B2}, which use self-expressing autoencoders to improve frame representations before segmenting and clustering, show higher boundary F-scores, comparable to the previous state of the art for systems privileging segmentation.

\section{Conclusion}

Major advances have been made towards unsupervised unit discovery for speech synthesis, at all levels---better units, better decoding architectures, and better waveform generation. The best discrete codes, however, are still  an order of magnitude more detailed than the phonemic representation. The supervised topline system demonstrates the possibility of a low bitrate code which is also of high quality. The challenge is to find such a high-quality low-bitrate phoneme-like representation in an unsupervised fashion. Nevertheless, some higher-bitrate codes may yet be useful, and good enough, to be used in language modelling. We will explore this in upcoming challenges.
Regarding spoken term discovery and segmentation, progress was made in this challenging dual task, with improved clusters and improved coverage. Clustering-oriented algorithms represent the best current tradeoff, but another potential path forward is to bring segmentation-oriented systems  towards better clusters. 

\section{Acknowledgements}
Funded by the ANR (ANR-17-EURE-0017 FRONTCOG, ANR-10-IDEX-0001-02 PSL*, ANR-18-IDEX-0001 U de Paris, ANR-19-P3IA-0001 PRAIRIE 3IA Institute,  ANR-17-CE28-0009 GEOMPHON, ANR-10-LABX-0083 EFL), CIFAR LMB, and a research gift by Facebook.



\let\oldbibliography\thebibliography
\renewcommand{\thebibliography}[1]{%
  \oldbibliography{#1}%
  \setlength{\itemsep}{0pt}%
}
\vspace{-5pt}

\bibliographystyle{IEEEtran}
\bibliography{mybib}

\clearpage
\setcounter{section}{0}
\setcounter{table}{0}
\setcounter{figure}{0}

\renewcommand\thesection{S\arabic{section}}
\renewcommand\thetable{S\arabic{table}}
\renewcommand\thefigure{S\arabic{figure}}

\end{document}